\pdfoutput=1

\documentclass[11pt]{article}

\usepackage{EMNLP2022}

\usepackage{times}
\usepackage{latexsym}
\usepackage{multirow}

\usepackage[T1]{fontenc}

\usepackage[utf8]{inputenc}

\usepackage{microtype}

\usepackage{inconsolata}

\usepackage{amsmath,amssymb}

\usepackage{algorithm} 
\usepackage{algorithmic}
\usepackage{graphicx}
\usepackage{arydshln}

\title{Multi-Granularity Optimization for Non-Autoregressive Translation}

\author{
 Yafu Li$^{\spadesuit \heartsuit}$\hspace{0.5mm}, 
 Leyang Cui$^{\clubsuit}$\footnotemark[2]\hspace{0.5mm}, 
 Yongjng Yin$^{\spadesuit \heartsuit}$\hspace{0.5mm}, 
 Yue Zhang$^{\heartsuit \diamondsuit}$\footnotemark[2]\hspace{0.2mm}\hspace{1.5mm} \\
 $^\spadesuit$ Zhejiang University\\
 $^\heartsuit$ School of Engineering, Westlake University\\
 $^\clubsuit$ Tencent AI lab \\
 $^\diamondsuit$ Institute of Advanced Technology, Westlake Institute for Advanced Study\\ 
 \textit{yafuly@gmail.com} \quad\textit{leyangcui@tencent.com}\\
 \textit{yinyongjing@westlake.edu.cn} \quad\textit{yue.zhang@wias.org.cn} \\
}
\date{}
\begin{document}
\maketitle

\renewcommand{\thefootnote}{\fnsymbol{footnote}}
\footnotetext[2]{Corresponding authors.}

\begin{abstract}

Despite low latency, non-autoregressive machine translation (NAT) suffers severe performance deterioration due to the naive independence assumption.
This assumption is further strengthened by cross-entropy loss, which encourages a strict match between the hypothesis and the reference token by token.
To alleviate this issue, we propose multi-granularity optimization for NAT, which collects model behaviors on translation segments of various granularities and integrates feedback for backpropagation.
Experiments on four WMT benchmarks show that the proposed method significantly outperforms the baseline models trained with cross-entropy loss, 
and achieves the best performance on WMT'16 En$\Leftrightarrow$Ro and highly competitive results on WMT'14 En$\Leftrightarrow$De for fully non-autoregressive translation.
\end{abstract}

\section{Introduction}

Neural machine translation (NMT) systems have shown superior performance on various benchmark datasets \cite{VaswaniSPUJGKP17,Edunov:emnlp18}. 
In the training stage, NMT systems minimize the token-level cross-entropy loss between the reference sequence and the model hypothesis. 
During inference, NMT models adopt autoregressive decoding, where the decoder generates the target sentence token by token ($O(N)$). 
To reduce the latency of NMT systems, \citet{nat} propose non-autoregressive neural machine translation (NAT), which improves the decoding speed by generating the entire target sequence in parallel ($O(1)$).

\begin{figure}[t]
\setlength{\belowcaptionskip}{0.5cm}
\centering
\includegraphics[width=0.95\linewidth]{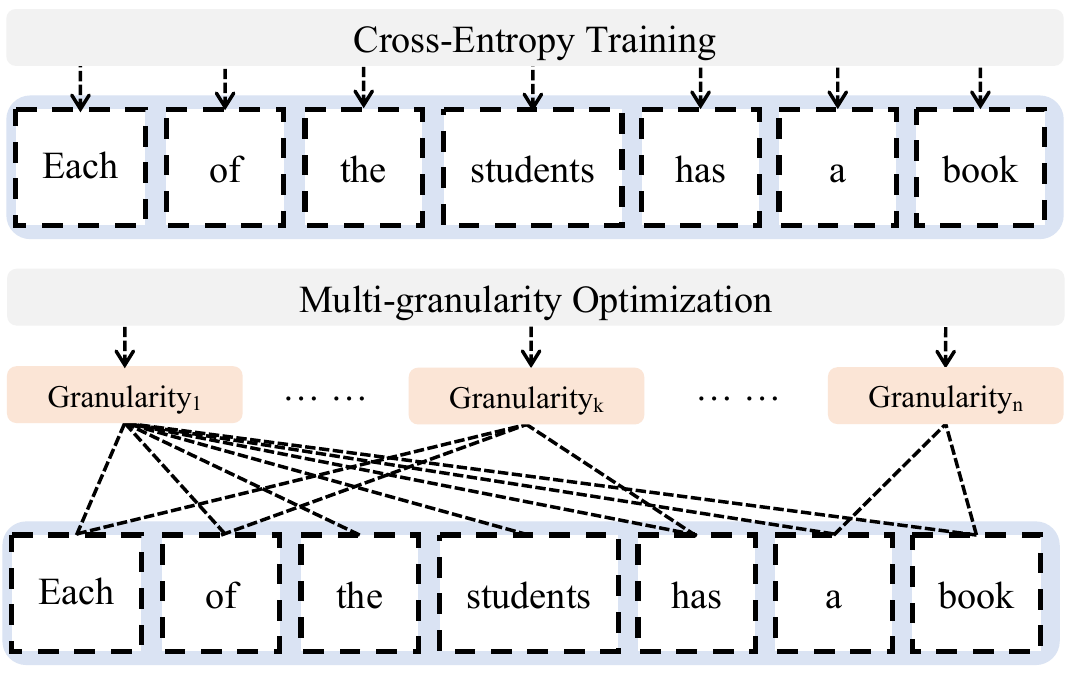}
\caption{
\label{fig:intro}
An illustration of modeling the multi-granularity token-dependency beyond cross-entropy.
}
\end{figure}

Despite low latency, without modeling the target sequence history, 
NAT models tend to generate translations of low quality \cite{nat, natcrf, maskp}. 
NAT ignores inter-token dependency and naively factorizes the sequence-level probability as a product of independent token probability.
However, vanilla NAT adopts the same training optimization method as autoregressive (AT) models, i.e., cross-entropy loss (XE loss), which forces the model to learn a strict position-to-position mapping, heavily penalizing hypotheses that suffer position shifts but share large similarity with the references. 
Given a reference ``she left her keys yesterday .'', an inappropriate hypothesis ``she left her her her .'' can yield a lower cross-entropy than one reasonable hypothesis ``yesterday she left her keys .''.
Autoregressive models suffer less from the issue by considering previous generated tokens during inference, which is however infeasible for parallel decoding under the independence assumption.
As a result, NAT models trained using cross-entropy loss are weak at handling multi-modality issues and prone to token repetition mistakes  \cite{natcrf, glat, axe}.



Intuitively, generating adequate and fluent translations involves resolving dependencies of various ranges \cite{yule2006study}.
For example, to generate a translation ``Each of the students has a book'', the model needs to consider the local n-gram pattern ``a - book'', the subject-verb agreement across the non-continuous span ``each - has'', and the global context. 
To capture the token dependency without the language model, 
feedback on model's behavior on text spans of multiple granularities can be incorporated.
To this end, we propose a multi-granularity optimization method to provide NAT models with rich feedback on various text spans involving multi-level dependencies.
As shown in Figure \ref{fig:intro}, instead of exerting a strict token-level supervision, we evaluate model behavior on various granularities before integrating scores of each granularity to optimize the model.
In this way, for each sample we highlight different parts of the translation, e.g., ``a book'' or ``each of the students has''.

\begin{figure*}[t]
\centering
\includegraphics[width=1\linewidth]{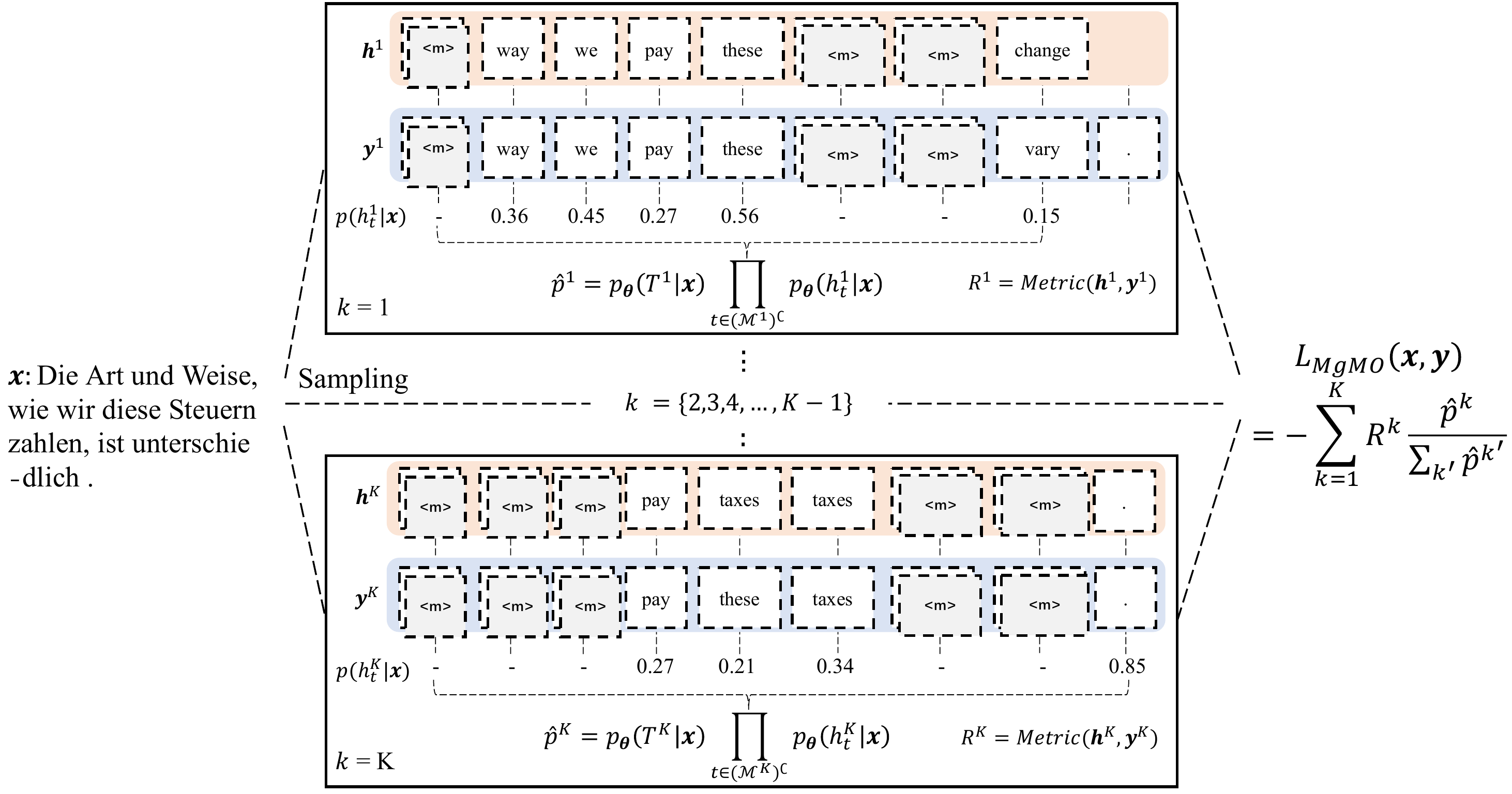}
\caption{
\label{method}
Method illustration of multi-granularity optimization for NAT. 
During training, our method (MgMO) samples $K$ hypotheses for each source sequence, and focuses on different parts of each one by applying the random masking strategy.
For example, MgMO collects model's performance on the partially exposed segments ``way we pay these'' and ``change'' for the first hypothesis ($\boldsymbol{h}^1$), while pays more attention to the phrase ``pay these taxes'' for the K-th one ($\boldsymbol{h}^K$).
}
\end{figure*}

During training, instead of searching for a single output for each source sequence, we explore the search space by sampling a set of hypotheses.
For each hypothesis, we jointly mask part of the tokens and those of the gold reference at the same positions.
To directly evaluate each partially masked hypothesis, we adopt metric-based optimization \cite{seq-rnn, mrt} which rewards the model with a metric function measuring hypothesis-reference text similarity.
Since both the hypothesis and the reference share the same masked positions, the metric score of each sample is mainly determined by those exposed segments.
Finally, we weigh each sample score by the model confidence to integrate the metric feedback on segments of various granularities. 
An illustrative representation is shown in Figure~\ref{method}, where a set of masked hypothesis-reference pairs are sampled and scored respectively before being merged by segment probabilities.
In this way, the model is optimized based on its behavior on text spans of multiple granularities for each training instance within a single forward-backward pass.


We evaluate the proposed method across four machine translation benchmarks: WMT14 En$\Leftrightarrow$De and WMT16 En$\Leftrightarrow$Ro. 
Results show that the proposed method outperforms baseline NAT models trained with XE loss by a large margin, while maintaining the same inference latency.
The proposed method achieves two best performances for fully non-autoregressive models among four benchmarks, and obtains highly competitive results compared with the AT model.
To the best of our knowledge, we are the first to leverage multi-granularity metric feedback for training NAT models. 
Our code is released at \href{https://github.com/yafuly/MGMO-NAT}{https://github.com/yafuly/MGMO-NAT}.


\label{sec:mgmo}

\section{Method}
We first briefly introduce some preliminaries including non-autoregressive machine translation (Section~\ref{sec:nat}) and cross-entropy (Section~\ref{sec:xe}),
and then we elaborate our proposed method where the model learns segments of different granularities for each instance (Section~\ref{sec:mgmo}).

\subsection{Non-autoregressive Machine Translation (NAT)}
\label{sec:nat}
The machine translation task can be formally defined as a sequence-to-sequence generation problem, where the model generates the target language sequence $\boldsymbol{y} = \{y_1,y_2,...,y_T\}$ given the source language sequence ${\boldsymbol{x} = \{x_1,x_2,...,x_S\}}$ based on the conditional probability $p_{\boldsymbol \theta}(\boldsymbol{y}|\boldsymbol{x})$ ($\boldsymbol \theta$ denotes the model parameters).
Autoregressive neural machine translation factorizes the conditional probability to: $\prod_{t=1}^{T}p(y_t|y_1,...,y_{t-1},\boldsymbol{x})$.
In contrast, non-autoregressive machine translation \cite{nat} ignores the dependency between target tokens and factorizes the probability as $\prod_{t=1}^{T}p(y_t|\boldsymbol{x})$, where tokens at each time step are predicted independently.


\subsection{Cross Entropy (XE)}
\label{sec:xe}
Similar to AT models, vanilla NAT models are typically trained using the cross entropy loss:
\begin{equation}
    \mathcal{L}_{XE} = - logp(\boldsymbol{y}|\boldsymbol{x}) = -\sum_{t=1}^{T}logp_{\boldsymbol{\theta}}(y_t|\boldsymbol{x})
\end{equation}

In addition, a loss for length prediction during inference is introduced:
\begin{equation}
    \mathcal{L}_{length} = -logp_{\boldsymbol{\theta}}(T|\boldsymbol{x})
\end{equation}

\citet{maskp} adopt masking scheme in masked language models and train the NAT models as a conditional masked language model (CMLM):
\begin{equation}
\label{eq_cmlm}
    \mathcal{L}_{CMLM} = -\sum_{y_t\in \mathcal{Y}(\boldsymbol{y})}logp_{\boldsymbol{\theta}}(y_t|\Omega(\boldsymbol{y},\mathcal{Y}(\boldsymbol{y})),\boldsymbol{x})
\end{equation}
where $\mathcal{Y}(\boldsymbol{y})$ is a randomly selected subset of target tokens and $\Omega$ denotes a function that masks a selected set of tokens in $\mathcal{Y}(\boldsymbol{y})$.
During decoding, the CMLM models can generate target language sequences via iteratively refining translations from previous iterations.


\subsection{Multi-granularity Optimization for NAT}
\label{sec:mgmo}
We propose multi-granularity optimization which integrates feedback on various types of granularities for each training instance.
The overall method illustration is presented in Figure \ref{method}.

\paragraph{Sequence Decomposition}
In order to obtain output spans of multiple granularities,
we sample $K$ output sequences from the model following a two-step sampling process.
In particular, we first sample a hypothesis length and then sample the output token at each time step independently given the sequence length. 
The probability of the $k$-th hypothesis $\boldsymbol{h}^k$ is calculated as:
\begin{equation}
    p_{\boldsymbol \theta}(\boldsymbol{h}^k|\boldsymbol{x}) = p_{\boldsymbol \theta}(T^k|\boldsymbol{x}) \prod_{t=1}^{T}p_{\boldsymbol \theta}(h_t^k|\boldsymbol{x})
\end{equation}



To highlight different segments of multiple granularities for each sample, we apply a masking strategy that randomly masks a subset of the tokens for both the hypothesis and the reference at the same position. 
We denote the masked hypothesis and reference as $\boldsymbol{h}^k = \{h^k_1,\dots,h^k_{T^k}\}$ and $\boldsymbol{y}^k=\{y^k_1,\dots,y^k_{T}\}$, respectively, and denote the set of masked positions as $\mathcal{M}^k$. 
Note that the reference length $T$ may be different from the hypothesis length $T^k$.

For the first hypothesis output ($k=1$) in Figure \ref{method}, given the randomly generated masked position set $\mathcal{M}^1=\{1,6,7\}$, the masked hypothesis and reference are $\boldsymbol{h}^1=\{\langle m \rangle ,h_2^{1},h_3^{1},...,\langle m \rangle,\langle m \rangle,h_8^{1}\}$ and $\boldsymbol{y}^1=\{\langle m \rangle,y_2^{1},y_3^{1},...,\langle m \rangle,\langle m \rangle,y_8^{1},y_9^{1}\}$, respectively, where $\langle m \rangle$ represents the masked token.
To determine the number of masked tokens $|\mathcal{M}^k|$ for each training instance, we first sample a threshold $\tau$ from a uniform distribution $\mathcal U(0,\gamma)$, and computes $|\mathcal{M}^k|$ as follows:
\begin{equation}
\label{eq_mask}
    |\mathcal{M}^k| = max( \lfloor T^k - \tau * T^k \rfloor,0)
\end{equation}
where $\gamma$ is a scaling ratio that controls the likelihood of being masked for each token. 
Note that the value of $|\mathcal{M}^k|$ lies within the range $[0,T^k-1]$, meaning that at least one token is kept.

In this way, we decompose each training instance into $K$ pairs of masked hypotheses and references with different granularities exposed.
For example, in the last sample ($k=K$) in Figure \ref{method}, only the verb phrase ``\textit{pay these taxes}'' and the period (``.'') are learned by the model, whereas the sample ($k=1$) reveals more informative segments.

\paragraph{Metric-based Optimization (MO)}
To avoid the strict mapping of XE loss \cite{axe, oaxe}, we incorporate metric-based optimization which is widely studied in reinforcement learning for NMT \cite{seq-rnn, mrt, revisit, study}.
Metric-based optimization allows more flexibility of token positions by rewarding the model with global scores instead of forcing it to fit a strict position-to-position mapping under XE.


The objective of metric-based optimization is to maximize the expected reward
with respect to the posterior distribution given the parameters $\boldsymbol \theta$. 
The reward $\Re(\boldsymbol \theta)$ for a training instance can be formally written as:
\begin{equation}
\label{eq_mgmo1}
    \Re(\boldsymbol \theta)=\sum_{\boldsymbol{h}\in \mathcal{H}(\boldsymbol{x})}p_{\boldsymbol \theta}(\boldsymbol{h}|\boldsymbol{x})R(\boldsymbol{h},\boldsymbol{y})
\end{equation}
where $\mathcal{H}(\boldsymbol{x})$ denotes the set of all possible candidate hypotheses for the source sequence $\boldsymbol{x}$. $R(\boldsymbol{h},\boldsymbol{y})$ denotes the metric function that measures the similarity between the hypothesis and the reference under a specific evaluation metric, e.g.,
GLEU \cite{gleu}.

\paragraph{Multi-granularity Metric-based Optimization (MgMO)}
Despite a lower gap between training and evaluation under the metric-based optimization, it suffers relatively coarse training signals as rewards are obtained by measuring sequence-level similarity.
However, due to lack of explicit token dependency, NAT requires more fine-grained feedback for capturing complex token dependencies that can spread across various lengths, e.g., continuous local dependencies and subject-verb agreement across the non-continuous spans.
We enrich the metric feedback in Equation \ref{eq_mgmo1} by decomposing a single sequence-level reward into multi-granularity evaluation.
For each training instance, the model is optimized by integrated feedback on its performance on various sequence segments of diverse granularities.

As enumerating segments of all possible granularities is intractable, we consider a finite set of sampled hypotheses to traverse as many granularities as possible.
Combining with sequence decomposition, we can rewrite the Equation \ref{eq_mgmo1} into a form of loss function:
\begin{equation}
\label{eq_rl}
    \mathcal{L}(\boldsymbol \theta)=-\sum_{\boldsymbol{h}^{k}\in \mathcal{K}(\boldsymbol{x})}p_{\boldsymbol \theta}(\boldsymbol{h}^k|\boldsymbol{x})R(\boldsymbol{h}^k,\boldsymbol{y}^k)
\end{equation}
where $\mathcal{K}(\boldsymbol{x})$ is a sampled subset consisting of $K$ sampled hypotheses. Applying the log derivative trick, the gradient can be derived:
\begin{equation}
    \frac{\partial \mathcal{L}(\boldsymbol \theta)}{\partial \boldsymbol \theta} = -\sum_{\boldsymbol{h}^{k}\in \mathcal{K}(\boldsymbol{x})}[R(\boldsymbol{h}^k,\boldsymbol{y}^k)\nabla_{\boldsymbol{\theta}}logp_{\boldsymbol{\theta}}(\boldsymbol{h}^k|\boldsymbol{x})]
\end{equation}
which does not require differentiation of the metric function $R$.
Since both the hypothesis and the reference are partially masked at the same positions, the exposed segments exert much larger effects on the metric function $R(\boldsymbol{h}^k,\boldsymbol{y}^k)$, i.e., the dissimilarity only origins from the unmasked tokens.

In order to make the model focus on the exposed segments, we transform the sequence probability in Equation \ref{eq_rl} into the probability of the segment co-occurrence. 
Since the output distributions of tokens are independent with each other in NAT, the segment co-occurrence probability is simply the multiplication of the probability of each unmasked token:
\begin{equation}
    \hat{p}_{\boldsymbol{\theta}}(\boldsymbol{h}^k|\boldsymbol{x}) = p_{\boldsymbol \theta}(T^k|\boldsymbol{x})\prod_{t\in (\mathcal{M}^k)^\complement}p_{\boldsymbol \theta}(h_{t}^{k}|\boldsymbol{x})
\end{equation}
where $(\mathcal{M}^k)^\complement$ denotes the complementary set of the masked position set $\mathcal{M}^k$.

Within the sample space, the sequence probability can be renormalized by model confidence \cite{Och03,mrt}:
\begin{equation}
    q_{\boldsymbol{\theta}}(\boldsymbol{h}^k|\boldsymbol{x};\alpha) = \frac{\hat{p}_{\boldsymbol \theta}(\boldsymbol{h}^k|\boldsymbol{x})^{\alpha}}{\sum_{\boldsymbol{h^{\prime}}\in \mathcal{K}(\boldsymbol{x})}\hat{p}_{\boldsymbol{\theta}}(\boldsymbol{h^{\prime}}|\boldsymbol{x})^\alpha}
\end{equation}
where $\alpha$ controls the distribution sharpness.

Taking the renormalized probability into Equation \ref{eq_rl}, the loss of multi-granularity metric optimization is formally written as:
\begin{equation}
\label{eq_loss_mgmo}
    \mathcal{L}_{MgMO}(\boldsymbol \theta)=-\sum_{k=1}^{K}q_{\boldsymbol \theta}(\boldsymbol{h}^k|\boldsymbol{x})R(\boldsymbol{h}^k,\boldsymbol{y}^k)
\end{equation}

In this way, MgMO decomposes the sequence-level feedback into pieces of multi-granularity metric feedback before integrating them to optimize the model, resulting in a set of more fine-grained training signals that examines different parts of the hypothesis.

\paragraph{Training}
\label{sec:train}
Following previous work \cite{seq-rnn, mrt, natbow, 
adeq, oaxe}, 
we adopt a two-stage training strategy, where CMLM loss is first applied for initialization and then replaced by MgMO loss for finetuning.
The length loss is maintained throughout the training process.

\begin{table*}[h!]
\small
    \centering
    \begin{tabular}{ccccc}
    \hline
         \multirow{2}{*}{\bf Model} & \multicolumn{2}{c}{WMT14} & \multicolumn{2}{c}{WMT16} \\
         & En-De & De-En & En-Ro & Ro-En \\
    \hline
         Autoregressive Transformer &  27.5 & 31.2 & 33.7  & 34.1 \\
         \hline
         NAT w/ Fertility \cite{nat}  & 17.6 & 19.8 & 24.5 & 25.7 \\
         CMLM ~\cite{maskp} & 18.3 & 22.0 & 27.6 & 28.6 \\
         Bag-of-ngrams Loss ~\cite{natbow}  & 20.9 & 24.6 & 28.3 & 29.3 \\
         Flowseq ~\cite{flowseq} & 21.5 & 26.2 & 29.3 & 30.4 \\
         Bigram CRF ~\cite{natcrf} & 23.4 & 27.2 & - & - \\
         CMLM + AXE ~\cite{axe} & 23.5 & 27.9 & 30.8 & 31.5 \\
         DSLP \& MT ~\cite{dslp} & 24.2 & 28.6 & 31.5 & 32.6 \\
         CMLM + EISL ~\cite{edit-inv} & 24.2 & - & - & - \\
         EM+ODD ~\cite{em} & 24.5 & 27.9 & - & - \\
         GLAT ~\cite{glat} & 25.2 & 29.8 & 31.2 & 32.0 \\
         Imputer ~\cite{imputer} & 25.8 & 28.4 & 32.3 & 31.7 \\
         CMLM + Order-Agnostic XE ~\cite{oaxe} & 26.1 & 30.2 & 32.4 & 33.3 \\ 
         AlignNART ~\cite{alignnart} & \textbf{26.4} & \textbf{30.4} & 32.5 & 33.1  \\
         latentGLAT ~\cite{latentglat} & \textbf{26.6} & 29.9 & 32.5 & - \\
         \hdashline
         CMLM$_{1}$ (our implementation) & 20.2 & 24.5 & 27.5 & 29.0 \\
         Metric-based Optimization (MO) & 24.8 & 29.1 & 31.2 & 32.4  \\
         Multi-granularity Metric-based Optimization (MgMO) & \textbf{26.4} & \textbf{30.3} & \textbf{32.9} & \textbf{33.6} \\
         \hline
    \end{tabular}
    \caption{Performance (test set BLEU) comparison of our proposed method (MgMO) with other fully non-autoregressive models (i.e., one-decoding pass). MgMO is significantly better than both the metric-based model (MO) and the CMLM baseline with $p<0.01$ \cite{sigtest}. Performance of other NAT models are obtained from corresponding papers. CMLM$_{1}$ denotes one-step decoding.} 
    \label{tab:main}
\end{table*}


\section{Experiments}
\subsection{Settings}
\paragraph{Data}
We conduct experiments on both directions of two standard machine translation datasets including WMT14 En$\Leftrightarrow$De and WMT16 En$\Leftrightarrow$Ro.
Knowledge distillation is commonly used for training NAT models \cite{nat, natcrf, maskp, axe}. We use the distilled dataset released by \citet{dslp} to obtain comparable baseline models.


\paragraph{Initialization}
We mainly follow \citet{dslp} for the model configuration.
We use Transformer and adopt Transformer\_Base configuration for all experiments: both the encoder and decoder consist of 6 layers with 8 attention heads, and the hidden dimension and feedforward layer dimension is 512 and 2,048, respectively.
We train the model using Adam \cite{adam} optimizer.
We set the weight decay as 0.01 and label smoothing as 0.1.
The learning rate increases to $5\cdot10^{-4}$ in the first 10K steps and then anneals exponentially. 
For WMT16 En$\Leftrightarrow$Ro, we use a dropout rate of 0.3 and a batch size of 32K tokens, whereas for WMT14 En$\Leftrightarrow$De, we switch to 0.1 and 128K accordingly.
Code implementation is based on Fairseq \cite{fairseq}.
We train all models for 300,000 steps and select the checkpoint with the best validation performance for MgMO finetuning.

\paragraph{Finetuning}
We finetune all models for 100,000 steps and use a dropout rate of 0.1.
The batch size for WMT16 En$\Leftrightarrow$Ro and WMT14 En$\Leftrightarrow$De is 256 and 1,024, respectively.
We use a fixed learning rate of $2\cdot10^{-6}$ during finetuning.
We use GLEU \cite{gleu} score as the metric function and set the value of $\alpha$ in Q-distribution as 0.005.
Based on validation results, we use a maximum n-gram size of 6 for GLEU score, set the scaling ratio $\gamma$ as 8 and set the sample space size $K$ as 40. 
\paragraph{Evaluation}
We use BLEU \cite{bleu} for all directions. 
Similar to autoregressive settings, we use $l=5$ length candidates during inference \cite{axe, oaxe}.
We select the best checkpoint for evaluation based on validation BLEU scores.

\subsection{Main Results}
We compare our method with the autoregressive Transformer and other fully NAT baselines (i.e., one decoding pass).
The results are shown in Table \ref{tab:main}.
We can observe that applying metric-based optimization (MO) on the CMLM baseline brings an improvement of 4.1 BLEU scores on average.
By decomposing the metric feedback into multi-granularity levels, MgMO further obtains a significant improvement (with $p<0.01$~\cite{sigtest}), expanding the advantage to 5.5 BLEU scores on average.
Compared with other representative NAT baselines, MgMO achieves the best performance on WMT16 En$\Leftrightarrow$Ro and highly competitive performance on WMT14 En$\Leftrightarrow$De.
Since no modification is involved in model architecture and inference, MgMO achieves the same inference latency with the vanilla NAT model.

\subsection{Training Strategies}

In addition to the default setting, we consider three alternative training strategies.
Generally, these strategies differ in how much information is fed into the NAT decoder and how much information left requires the decoder to fit: 
(1) none of the target tokens are observed and the complete set of target tokens are considered for computing metric feedback (\textbf{N\&C});
(2) part of the target tokens are observed and the complete set of target tokens are considered for computing metric feedback (\textbf{P\&C}); 
(3) part of the target tokens are observed and a partial set of the target tokens are highlighted for computing metric feedback (\textbf{P\&P}); 
(4) none of the target tokens are observed and a partial set of the target tokens are highlighted for computing metric feedback (\textbf{N\&P}, i.e., the default setting). 
We present an example in Table \ref{tab:strategy_example} as an intuitive illustration for different training strategies. 
\begin{table}[t]
\centering
\small
\begin{tabular}{c c}
    \hline
    \multirow{2}{*}{\textbf{Strategy}} & \textbf{Decoder Input} \\
    & \textbf{Decoder Target}\\
     \hline
     \multirow{2}{*}{\textbf{N\&C}} &  <unk> <unk> <unk> <unk> <unk>\\
      & I never went back . \\
     \hline
     \multirow{2}{*}{\textbf{P\&C}} & <unk> <unk> went back <unk> \\
      & I never went back . \\
     \hline
     \multirow{2}{*}{\textbf{P\&P}} & <unk> <unk> went back <unk> \\
      & I never <m> <m> .\\
     \hline
     \multirow{2}{*}{\textbf{N\&P}} & <unk> <unk> <unk> <unk> <unk>\\
      & I never <m> <m> . \\
     \hline
\end{tabular}
    \caption{Sample decoder input-target pairs for the source German sentence ``Ich kehrte nie zurück .'', under different training strategies. For decoder input, ``<unk>'' denotes the unknown token in the dictionary and indicates null information. For decoder traget, segments that are not replaced by ``<m>'' are highlighted for computing metric feedback.}
    \label{tab:strategy_example}
\end{table}


 

\begin{table}[t]
\centering
\small
\begin{tabular}{c c c c c}
    \hline
     \multirow{2}{*}{\textbf{Dataset}} & \multicolumn{4}{c}{\textbf{Training Strategy}}\\
      &\textbf{N\&C} & \textbf{P\&C} & \textbf{P\&P} & \textbf{N\&P}\\
     \hline
    \textbf{WMT16 En$\Rightarrow$Ro} & 32.0 &  32.7 & 32.7 & \textbf{32.9} \\
     \textbf{WMT16 Ro$\Rightarrow$En} &  33.8 & 34.3 & 34.4 & \textbf{34.5} \\
     \hline
\end{tabular}
    \caption{Comparison of different training strategies on WMT16 En$\Leftrightarrow$Ro validation set.}
    \label{tab:strategy}
\end{table}
We conduct experiments on the WMT16 En$\Leftrightarrow$Ro dataset. 
The results are shown in Table \ref{tab:strategy}. 
We can observe that both partial observations (P\&C and P\&P) and partial predictions (P\&P and N\&P) obtain consistent improvement over the N\&C strategy, i.e., metric-based optimization without multi-granularity (MO in Table \ref{tab:main}).
P\&C achieves similar performance to P\&P, since it is easy for the model to copy the partial observations as the corresponding predictions, and thus focus on the unobserved tokens.
In other words, P\&C, P\&P and N\&P all incorporate training signals from multiple granularities, either explicitly or implicitly, and therefore obtain better performance.
Due to a smaller discrepancy between training and inference, N\&P obtains further performance improvement over P\&C and P\&P.

\begin{figure}[t]
\centering
\includegraphics[width=0.7\linewidth]{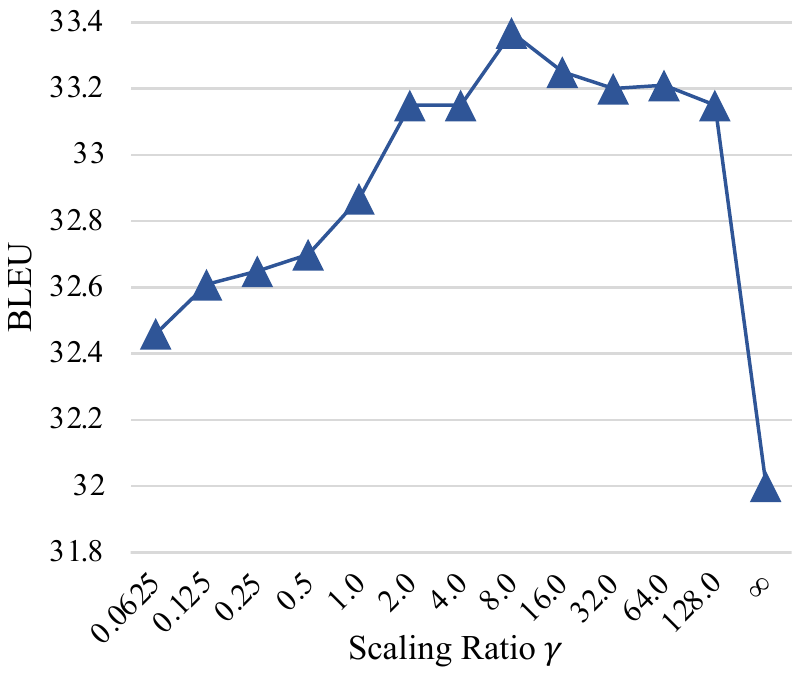}
\caption{
\label{fig:gamma}
Effect of $\gamma$ on the WMT En$\Rightarrow$Ro validation set. A larger $\gamma$ indicates tokens in hypotheses and corresponding references are less likely to be masked.
}
\end{figure}

\begin{figure}[t]
\centering
\includegraphics[width=0.9\linewidth]{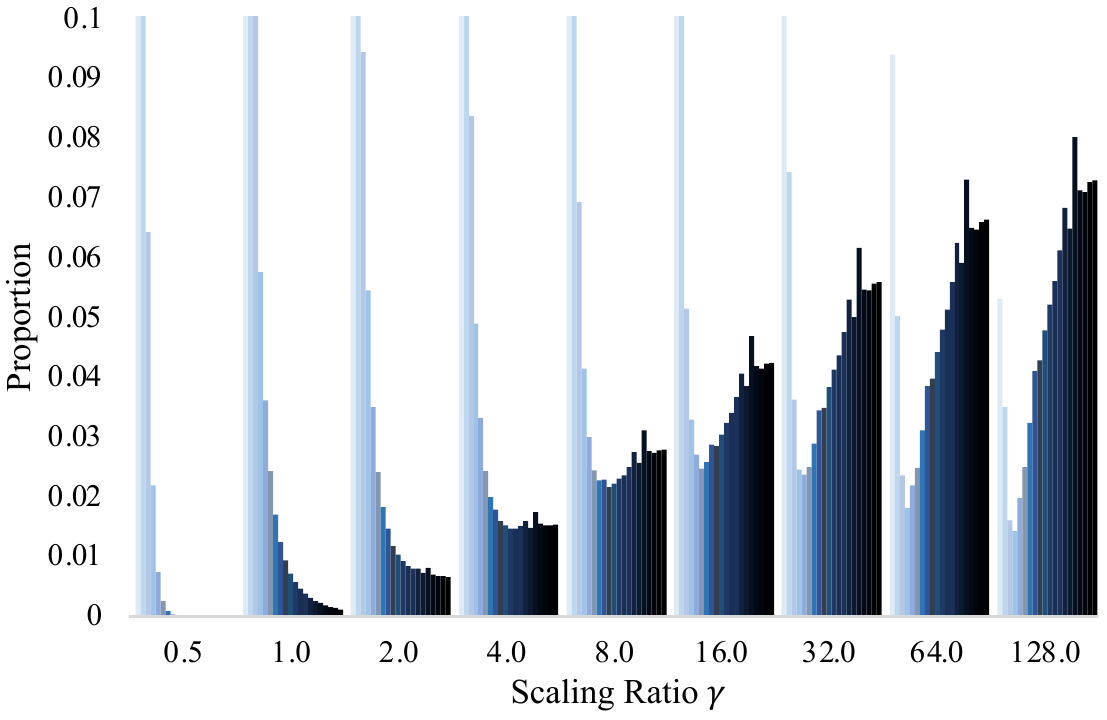}
\caption{
\label{fig:gran}
Proportions of granularities (1-gram to 20-gram) under different scaling ratios $\gamma$. Darker colors indicate larger granularities. 
}
\end{figure}

\begin{figure}[t]
\centering
\includegraphics[width=0.8\linewidth]{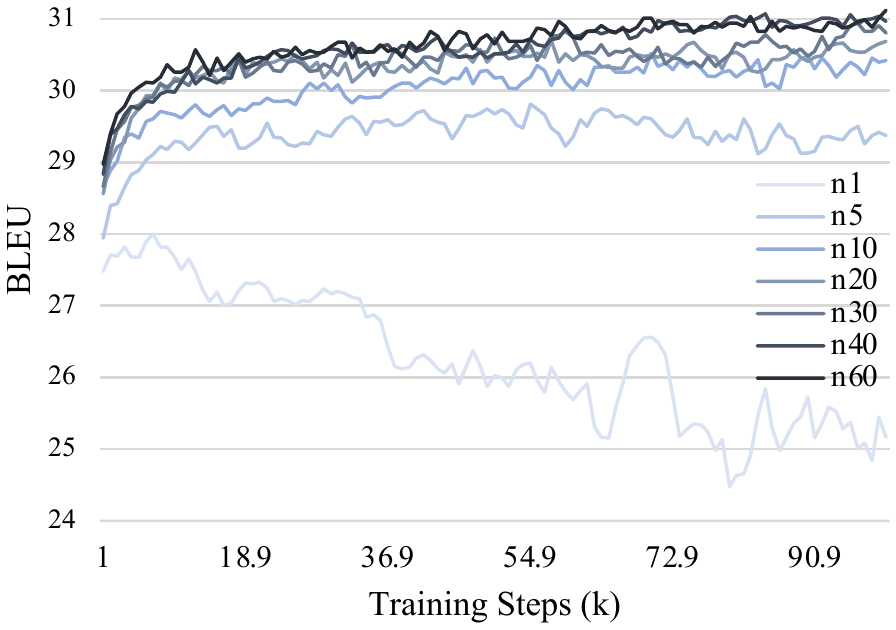}
\caption{
\label{fig:sample_space}
Effect of sample space size $K$ on the WMT En$\Rightarrow$Ro validation set. Darker colors indicate more samples are explored for each training instance.
}
\end{figure}
\begin{table}[t]
\centering
\small
\begin{tabular}{ c c c c c}
    \hline
    \multicolumn{5}{c}{\textbf{Maximum N-gram Size}} \\
\textbf{ 1-gram} & \textbf{2-gram} & \textbf{4-gram} & \textbf{6-gram}  \\
     \hline
    32.7 & 32.7 & \textbf{32.9} & \textbf{32.9}\\
     \hline
\end{tabular}
    \caption{Effect of maximum n-gram size of the GLEU metric on the WMT16 En$\Rightarrow$Ro validation set.}
    \label{tab:gram}
\end{table}

\begin{table}[t!]
\centering
\small
\begin{tabular}{ c c c c c c c}
    \hline
    \multicolumn{6}{c}{\textbf{Metric Function for Optimization}} \\
\textbf{BLEU} & \textbf{GLEU} & \textbf{chrF} & \textbf{TER} & \textbf{Rouge-2}  \\
     \hline
    32.7 & \textbf{32.9} &  \textbf{32.9} & 32.6 & 32.7 \\
     \hline
\end{tabular}
    \caption{Effect of different metric functions on the WMT16 En$\Rightarrow$Ro validation set.}
    \label{tab:metric}
\end{table}
\subsection{Ratio of Exposed Segments}
The scaling ratio $\gamma$ controls how likely one token is masked, i.e., a larger $\gamma$ indicates a higher probability of being exposed for each token and a lower probability otherwise. 
In general, a proper scaling ratio yields more diverse granularities from which the model learns rich token dependencies. 

As can be seen from the Figure \ref{fig:gamma}, increasing the scaling ratio $\gamma$ until 8.0 steadily brings performance improvement in terms of validation BLEU scores.
This can be because a larger $\gamma$ results in longer and more informative segments across different samples.
Therefore, the model is encouraged to learn to handle longer and more difficult token dependency, which is common in long sequences and is
a major challenge for NAT \cite{nat, glat, oaxe}.
Analysis of performance on different sequence lengths (Section~\ref{ana:length}) further validate our assumption.

As the ratio increases further, model performance begins to deteriorate since overly large ratios nearly expose all tokens for each sample, 
resulting in coarse granularities with limited diversity.
The extreme case
becomes the N\&C strategy ($\gamma\rightarrow\infty$), which reveals all target tokens in every sample.
In this case, the complete sequence becomes the only granularity, giving a coarse and monotonous feedback that is hard
to learn from.

To provide a statistical intuition of why the ratio of 8 obtains better performance, we traverse the training set (WMT'16 En$\Rightarrow$Ro) under different scaling ratios, and calculate the ratios of different granularities, i.e., the segments of different lengths.
The results are shown in Figure \ref{fig:gran}.
We can observe that masking with a larger ratio spreads more proportions on larger granularities.
While maintaining a significant portion on large granularities,
the ratio of 8.0 yields a relatively smooth distribution over various granularities, 
contributing to a progressive learning curriculum.



\subsection{Size of Search Space}

Intuitively, a larger sample space, i.e., a larger $K$, brings better granularity diversity since more sets of segments are encountered. 
On the other hand, enlarging the sample space increases computational complexity. 
We explore the effect of sample space size $K$, with results shown in Figure \ref{fig:sample_space}.
We can observe that increasing $K$ up to 40 brings a steady performance improvement on the validation set.
A larger $K$ (i.e., 60) does not lead to further improvement, indicating that the model has encountered sufficient types of granularities.

\subsection{Alternative Optimization Targets}
Beyond the standard GLEU metric, we also explore the effects of the maximum n-gram size and other alternative metrics for optimization.
\paragraph{Effects of N-gram Size}
We vary the maximum n-gram size in GLEU score and the results are shown in Table \ref{tab:gram}.
We can observe that rewarding local matches (1-gram and 2-gram) obtains comparable performance to that of larger span matches (4-gram and 6-gram).
We hypothesize that multi-granularity optimization compensates for capturing word ordering in some degree by simultaneously evaluating various granularities of the target sequence, which implicitly restricts the token locations.



\paragraph{Alternative Metric Functions}
We also explore the model performance under different metric functions, with results presented in Table \ref{tab:metric}.
We can see that different metric functions achieve comparable performance as they all measure sequence-level similarities.
Specifically, the metrics (GLEU and chrF \cite{chrf}) considering both n-gram precision and recall obtain better performance compared with BLEU which only considers precision.
Since Rouge-2 \cite{rouge} considers a maximum n-gram size of 2, it achieves worse performance than GLEU.
TER \cite{ter} measures the number of edit operations and 
obtains a slightly worse performance, since it does not reward n-gram match and suffers a discrepancy from the evaluation metric, i.e., BLEU.

\begin{figure*}[t]
\setlength{\belowcaptionskip}{-0.2cm}
\centering
\includegraphics[width=1\linewidth]{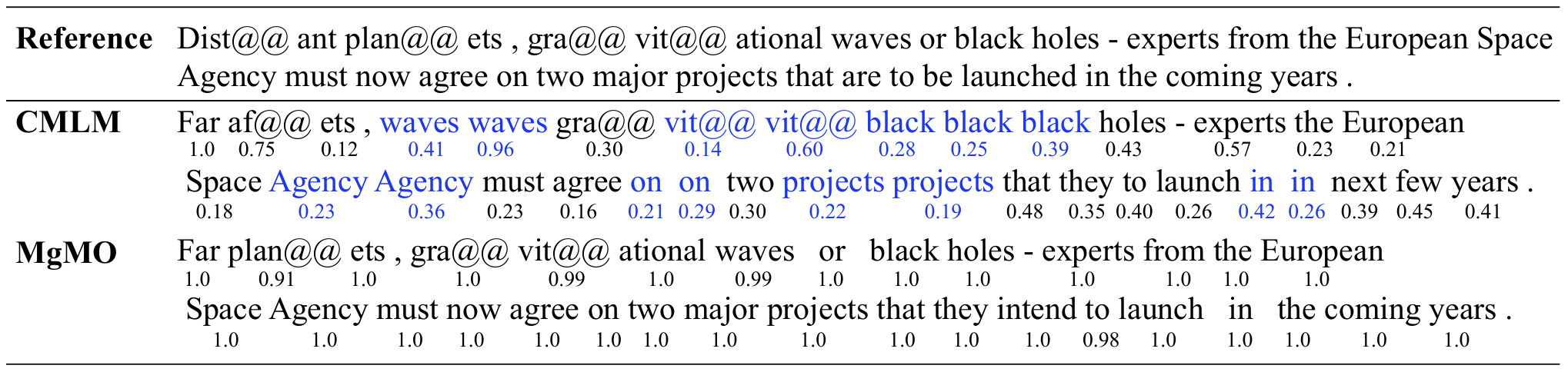}
\caption{
\label{fig:case}
An example from the WMT'14 De$\Rightarrow$En test set, with confidence (probability of generating the token) annotated below each token. \textcolor{blue}{Blue color} denotes token repetitions. }
\end{figure*}

\section{Analysis}
In this section, we conduct quantitative and qualitative analysis to dig some insights into how MgMO benefits non-autoregressive translation.
\subsection{Sequence Lengths}
\label{ana:length}
\begin{figure}[t]
\centering
\includegraphics[width=0.8\linewidth]{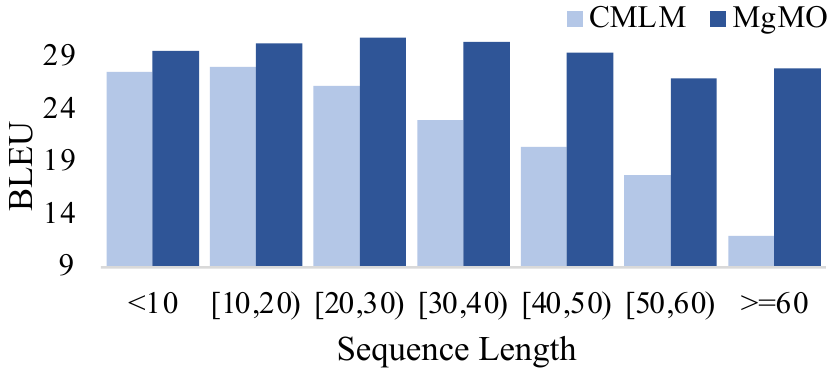}
\caption{
\label{fig:len}
Comparison of BLEU scores with respect to the lengths of the reference sentences on the WMT'14 De$\Rightarrow$En test set.
}
\end{figure}

We analyze the effectiveness of our method by comparing performance on test sentences of different lengths. 
We use compare-mt \cite{compare-mt} to split the WMT’14 De$\Rightarrow$En test sets into several subsets based on target sequence lengths.
The results are shown in Figure \ref{fig:len}.
As the sequence length grows, the baseline model trained using XE loss suffers great performance deterioration.
Under MgMO, the model maintains relatively stable performance on test sentences across different lengths, proving that multi-granularity learning brings benefits for capturing non-local dependencies that can spread across long text spans.

\begin{table}[t]
\centering
\small
\begin{tabular}{c c c }
    \hline
     \multirow{2}{*}{\textbf{Model}} & \multicolumn{2}{c}{\textbf{WMT'14}}\\
      &\textbf{En$\Rightarrow$De} & \textbf{De$\Rightarrow$En} \\
     \hline
      CMLM + OaXE &0.19&  0.19  \\
     \hdashline
    CMLM & 0.76 & 0.67 \\
    MgMO & \textbf{0.10} & \textbf{0.10}\textbf{} \\
     \hline
\end{tabular}
    \caption{Normalized Corpus-level multimodality (NCM) scores of the sentences on the WMT'14 En$\Leftrightarrow$De test set. The results of CMLM + OaXE are obtained from \citet{oaxe}. }
    \label{tab:ncm}
\end{table}

\subsection{Prediction Confidence}
NAT shows weakness in handling multi-modality \cite{nat, natcrf}, which is reflected by its low confidence on locating token translations among neighboring positions \cite{axe, oaxe}.
Ideally, we expect each token to have a high probability mass at the position it is predicted, but low at the neighboring positions.
Following previous work \cite{em,oaxe}, we compute Normalized Corpus-level multimodality (NCM) on the WMT14 En$\Leftrightarrow$De test set which measures average token-level prediction confidence.
The results are shown in Table \ref{tab:ncm}, and lower NCM scores indicate higher confidence.
We can see that applying MgMO largely increases the model prediction confidence at each step.
This can be because MgMO better captures token interdependency via optimizing model predictions based on various contexts, i.e., different sets of exposed segments.
We show an example in Figure~\ref{fig:case} to provide an intuition of effects brought by higher prediction confidence.

\section{Related Work}

\paragraph{Fully Non-Autoregressive Models}
To bridge the performance gap between fully NAT and the autoregressive counterpart, lots of techniques have been proposed to build dependencies among the target tokens such as
curriculum learning \cite{clnat,tclnat,mgclnat}, latent variable modeling \cite{ctc18,dislatent,flowseq,imputer,latentglat,tgtcodenat}, improving distillation training \cite{kdnat,lexinat,lowfwords}, and adaptive token sampling \cite{glat}.
Despite their success, these methods are trained with XE loss, which forces a strict mapping.
Another line of work shift to improvement of XE loss or metric-based objectives. 
For example, \citet{axe} soften the penalty for word order errors based on a monotonic alignment assumption, and \citet{oaxe} computes XE loss based on the best possible alignment between predictions and target tokens.
In contrast, we unitize hypothesis sampling and metric-based optimization, allowing the model to explore hypothesis of different lengths.
\citet{natcrf} incorporate an approximation of Conditional Random Fields (CRF) to model output dependency, while the decoding is not parallelized.
\citet{bleunat} devise customized reinforcement algorithms to optimize global metrics for NAT.
\citet{natbow} and \citet{edit-inv} propose differentiable n-gram matching losses between the hypothesis and reference.
In comparison, we propose to integrate feedback from evaluating model behavior on multi-level granularities within a single forward-backward propagation.

\paragraph{Metric-based Optimization for NMT}
Metric-based optimization has been utilized in NMT \cite{seq-rnn,mrt,acnmt,study,losses4seq,adeq,revisit} to alleviate the mismatch between the optimization during training and evaluation during inference.
For example, \citet{seq-rnn} train NMT using the objective gradually shifting from token-level likelihood to sentence-level BLEU score, and \citet{mrt} adopt minimum risk training (MRT) to minimize the task specific expected loss (i.e., induced by BLEU score). 
\citet{adeq} use translation adequacy as the metric function.
Different from them, we propose to integrate metric feedback on various granularities instead of a coarse sequence-level reward. 

\section{Conclusion}
We proposed multi-granularity optimization for NAT, which considers metric feedback on hypothesis segments of multiple granularities. 
Through integrating multi-granularity feedback, the model is optimized by focusing on different parts of the sequence within a single forward-backward pass, obtaining more detailed and informative training signals.
Empirical results demonstrated that our method achieved highly competitive performance compared with other representative baselines for fully NAT.
Analysis further showed that MgMO maintained strong performance on long sequences that vanilla NAT models suffer, and obtained high prediction confidence.
Beyond non-autoregressive translation, our proposed method can be used in other text generation tasks.
\section*{Limitations}
Despite competitive performance, the model still suffers common issues in NMT. 
Firstly, although the total GPU memory cost for our method is lower than that of XE loss (as the batch size under MgMO is much lower), MgMO requires a relatively large minimum memory capacity since lots of samples are considered for forward and backward propagation for each training instance. 
Secondly, like many other NAT models, our model also suffers, though smaller, performance deterioration without using data distilled from autoregressive teacher models.
\section*{Acknowledgment}
We thank all reviewers for their insightful comments. 
This publication has emanated from research conducted with the financial support of the Pioneer and "Leading Goose" R\&D Program of Zhejiang under Grant Number 2022SDXHDX0003.
This work is also under a grant from Lan-bridge Information Technology Co., Ltd. 
\bibliography{anthology,final}
\bibliographystyle{acl_natbib}


\end{document}